\documentclass[12pt,doublecolumn]{IEEEtran}

\usepackage{balance}  

\usepackage{booktabs}
\usepackage{fancyhdr}
\usepackage{amssymb}
\usepackage{amsmath}
\usepackage{comment}
\usepackage{algorithm}
\usepackage{algorithmic}
\usepackage{amsmath}
\usepackage{graphicx}
\usepackage{wrapfig}
\usepackage{url}
\usepackage{subcaption}
\def\i#1{\hbox{\it #1\/}}
\def\beq{\begin{equation}}
\def\eeq#1{\label{#1}\end{equation}}
\def\ba{\begin{array}}
\def\ea{\end{array}}

\def\iif{\hbox{\bf if}}

\def\causes{\hbox{\bf causes}}

\begin{document}

\title{Interpretable Automated Machine Learning \\ in Maana$^{\scriptsize\hbox{TM}}$ Knowledge Platform}

%
%

\author{Alexander Elkholy, Fangkai Yang, Steven Gustafson \\
Maana, Inc.\\
Belleue, Wa.\\
aelkholy@maana.io
}

\maketitle


\begin{abstract}
Machine learning is becoming an essential part of developing solutions for many industrial applications, but the lack of interpretability hinders wide industry adoption to rapidly build, test, deploy and validate machine learning models, in the sense that the insight of developing machine learning solutions are not structurally encoded, justified and transferred. In this paper we describe Maana Meta-learning Service, an interpretable and interactive automated machine learning service residing in Maana Knowledge Platform that performs machine-guided, user assisted pipeline search and hyper-parameter tuning and generates structured knowledge about decisions for pipeline profiling and selection. The service is shipped with Maana Knowledge Platform and is validated using benchmark dataset. Furthermore, its capability of deriving knowledge from pipeline search facilitates various inference tasks and transferring to similar data science projects.
\end{abstract}

\section{Introduction}
Machine learning is becoming an essential part of developing solutions for many industrial applications. Developers of such applications need to rapidly build, test, deploy and validate machine learning models. The validation of models is a key capability that will enable industries to more widely adopt machine learning capabilities for business decision making, however, this process suffers from the lack of {\em interpretability}. First, in most cases, validation begins with understanding how a machine learning model is developed - the pipeline from source data, through data processing and featurization, to model building and parameter tuning.  The capability of understanding machine learning models also represents a key piece of domain knowledge: data scientists who understand how to make successful domain specific machine learning pipelines will be in high demand across that domain.  Unfortunately, this level of understanding remains somewhat of a "dark art" in that the knowledge and judgment used to find good domain-specific machine learning pipelines is usually found in the heads of the data scientists.  Therefore, while it is possible to see the final machine learning pipeline, the steps the data scientist went through, and the compromises and decisions they made, are not captured. Once the model is delivered, most insights and assumptions related to development of the solution are lost, making long term sustaining difficult. Second, since there is no clear way of encoding the empirical experience of the data scientist derived from developing data science solutions to facilitate knowledge transfer and sharing so that they can be applied to similar projects efficiently and shared among a group of data scientists in the organization, it causes repetitive work, low efficiency and inconsistency in the quality of solutions.

To facilitate rapid development and deployment of data science solutions, automated machine learning (AutoML) has gained more interest recently due to the availability of public dataset repositories and open source machine learning code bases.  AutoML systems such as Auto-WEKA \cite{ThoHutHooLey13-AutoWEKA}, Auto-SKLEARN \cite{feurer2015efficient} and TPOT\cite{olson2016evaluation} attempt to optimize the entire machine learning pipeline, which can consist of independent steps such as featurization, which encodes data in numeric form, or feature selection, which attempts to pick the best subset of features to train a model from data. Given sufficient computation resources, these system can achieve good accuracy in building machine learning pipelines, but they do not provide clear explanations to justify the choice of models that can be verified by data scientist, and consequently the problem of interpretability remains unsolved.

To address this challenge, in this paper we describe the \textbf{Maana Meta-learning} service which provides {\em interpretable automated machine learning}. The goal of this project is two-folded. First, we hope that the efficiency of developing data science solutions can be improved by leveraging an automated search and profiling algorithm such that a baseline solution can be automatically generated for the data scientists to fine-tune. Second, we hope that such automated search process is transparent to human users, and through learning process the service can return interpretable insights on the choice of models and hyper-parameters and encode them as knowledge. Contrasted with most AutoML systems that provide end-to-end solutions, the Maana Meta-learning service is an interactive assistant to data scientists that performs {\em user-guided, machine-assisted} automated machine learning. By having data scientists specify a pre-determined search space, and Meta-learning service then goes through several stages to perform model selection, pipeline profiling and hyper-parameter tuning. During this process, it returns intermediate results and user can inject feedback to steer the search process. Finally, it generates an optimal pipeline along with structured knowledge encoding the decision making process, leading to an interpretable automated machine learning process.

Maana Meta-Learning service features two components: (1) a knowledge representation that captures domain knowledge of data scientists and (2) an AutoML algorithm that generates machine learning pipeline, evaluates their efficacy by sampling hyper-parameters, and encodes all the information about the choices made and subsequent performance / parameters into the knowledge representation. The knowledge representation is defined using GraphQL\footnote{https://graphql.org/}. Developed by Facebook as an alternative to the popular REST interface \cite{fielding2000architectural}, GraphQL provides only a single API endpoint for data access, backed by a structured, hierarchical type system. Consequently, it allows us to define a knowledge taxonomy to capture concepts of machine learning pipelines, seamlessly populate facts to the predefined knowledge graph and reason with them. The AutoML algorithm, in charge of generating and choosing which pipelines to pursue, is based on PEORL framework \cite{yang:peorl:2018}, an integration of symbolic planning \cite{cim08} and hierarchical reinforcement learning \cite{sutton-barto:book}. Symbolic plans generated from a pre-defined symbolic formulation of a dynamic domain is used to guide reinforcement learning, and recently this approach is generalized to improve interpretability of deep reinforcement learning. In the setting of AutoML, generating machine learning pipelines is treated as a symbolic planning problem on an action description in action language $\mathcal{BC}$\cite{lee13} that contains actions such as {\em preprocessing}, {\em featurizing}, {\em cross validation}, {\em training} and {\em prediction}.  
The pipeline is sent to execution where each symbolic action by mapping to primitive actions in a Markov Decision Process \cite{puterman} (MDP) space, which are ML pipeline components instantiated with random hyper-parameters, in order to learn the quality of the actions in the pipeline. The learning process is value iteration on R-learning \cite{rlearning:schwartz,sm:mlj96}, where cross-validation accuracy of the pipeline is used as rewards. After the quality of the current pipeline is measured, an improved ML pipeline is generated thereafter using the learned values, and the interaction with learning continues, until no better pipeline can be found. This step is called {\em model profiling}. After that, a more systematic parameter sweeping is performed, i.e., {\em model searching}. This allows us to describe the pipeline steps in an intuitive representation and explore the program space more systematically and efficiently with the help of reinforcement learning.

In this paper, we demonstrate that Maana Meta-learning provides a decent baseline on a variety of data sets, involving both binomial and multinomial classification tasks on various data types. Furthermore, when knowledge instance is filled into the pre-defined knowledge schema, the insights derived from Meta-learning process can be visualized as a knowledge graph, improving interpretability and facilitating knowledge sharing, sustaining as well as transferring to similar tasks. We show that by using an interactive process leveraging domain knowledge and user feedback to populate knowledge into a structured knowledge graph, in order to address the interpretable automated machine learning sought after by industrial application of data science.


%
%

\section{Related Work}
In contrast with optimizing the selection of model parameters, the goal of the AutoML task is to optimize an entire machine learning \textit{pipeline}. That is, starting from the raw data, it concerns itself with everything, including optimal selection of featurization, selection of algorithm, hyper-parameter selection as well as the cohesive collection of these as an ensemble. The most recent and most relevant approaches to the AutoML paradigm are Auto-WEKA and Auto-SKLEARN. Auto-WEKA \cite{ThoHutHooLey13-AutoWEKA} \cite{kotthoff2017auto} call this the \textit{combined algorithm selection and hyperparameter optimization problem} or CASH. This approach is formalized as a Bayesian optimization problem where it sequentially tests different pipelines based on the performance of the last using what is called a sequential model-based optimization (SMBO) formulation \cite{hutter2011sequential}. In combination with the algorithms available in the WEKA library\cite{eibe2016weka}, they provide a complete package targeted toward non-expert users, allowing them to build machine learning solutions without necessarily knowing the details required to do so. Auto-SKLEARN \cite{feurer2015efficient} approaches the AutoML task in much the same way by treating it as a Bayesian optimization problem. However, they claim that by giving a "warmstart" to the optimization procedure, the time to reach performant pipelines is significantly reduced. That is, they pre-select possible good configurations to begin the procedure. Thus their goal is to increase the efficiency of and reduce time to build. Additionally, instead of the WEKA library, they use the scikit-learn library \cite{scikit-learn}. Recent system KeyStoneML \cite{sparks2017keystoneml} uses technique similar to database query optimization to optimize machine learning pipelines end-to-end, where ML operator has a declarative logical representation.
By comparison, our work has a different focus and scope. Instead of directly outputting the best machine learning pipeline and providing a one-to-one solution, we focus on an interactive process where data scientists use this service to explore their predefined search space and refine their decision. In this setting, Meta-learning provides ``user-guided, machine assisted" automated search and facilitates encoding knowledge and decision making process and address the challenge of interpretability of data science solutions. The intepretability of automated machine learning and knowledge derived from the search algorithm is enabled leveraging PEORL framework \cite{yang:peorl:2018}, which is a combination of symbolic planning and reinforcement learning. Symbolic planning \cite{cim08} generates possible sequences of actions to achieve a goal which are pre-defined and logic-based (e.g. only certain data types are compatible with certain featurizers). Combined with R-Learning \cite{rlearning:schwartz}, feedback on actions taken is learned in order to generate new plans.

%
%

\section{Preliminaries}
\subsection{GraphQL and Maana Knowledge Platform}
GraphQL is a unified layer for data access and manipulation. In a distributed system, it is located at the same layer like REST, SOAP, and XML–RPC, that means it is used as an abstraction layer to hide the database internals. A GraphQL schema consists of a hierarchical definition of types and the operations that can be applied on times, i.e., queries and mutations. GraphQL’s type system is very expressive and supports features like inheritance, interfaces, lists, custom types, enumerated types. By default, every type is nullable, i.e. not every value specified in the type system or query has to be provided. Every GraphQL type system must specify a special root type called Query, which serves as the entry point for the query’s validation and execution. One example of a GraphQL schema definition is shown as follows. It contains two types: \texttt{Person} that contains fields of name (the punctual \texttt{!} denotes non-empty fields), age, a list of instances of books (denoted by brackets \texttt{[]}), and a list of instances friends, and a type Book with fields title and a list of persons as authors. Furthermore, there are 3 queries that retrieve an instance of Person by name, an instance of book by title, and a list of books by applying a filter. There is also a mutation that adds a person by providing a name.
{\small
\begin{verbatim}
type Person{
    name : String!
    age : Integer
    books(favorite: Boolean) : [Book]
    friends : [Person]
}
type Book {
    title : String!
    authors : [Person]
}
type Query {
    person(name : String!) : Person
    book(title : String!) : Book
    books(filter : String!) : [Book]
}
type Mutation {
    addPerson(name:String!) : Boolean
}
\end{verbatim}}
Such schema provides an representational abstraction of operations and the data they manipulates, and connects the front-end query/mutation calls with the back-end implementation details. 

Maana Knowledge Platform\footnote{https://www.Maana.io/knowledge-platform/} is architected based on Graphql-based microservices where their type systems are connected with each other to become a {\em Computational Knowledge Graph} (CKG). Different from traditional semantic systems based on ontology and description logic \cite{baader2003description}, the CKG separates the conceptual modeling of data, the content of the data and the operations on the data. This separation enables a fluidity of modeling, allowing data from any source and in any format to be seamlessly integrated, modeled, searched, analyzed, operationalized and re-purposed. Each resulting model is a unique combination of three key components – subject-matter expertise, relevant data from silos, and the right algorithm – all of which are instrumental in optimizing operations and decision flows. Furthermore, the CKG is also dynamic, which means that it can represent conceptual and computational models. In addition, it can be used to perform complex transformations and calculations at interactive speeds, making it a game-changing technology for agile development of AI-driven knowledge applications.

\subsection{PEORL Framework}
PEORL \cite{yang:peorl:2018} is a framework that integrates symbolic planning with reinforcement learning \cite{sutton-barto:book}. Using a symbolic formulation to capture high-level domain dynamics and planning with it, a symbolic plan is used to guide reinforcement learning to explore the domain, instead of performing random trial-and-error. Due to the fact that domain knowledge significantly reduces the search space, this approach accelerate learning and also improves the robustness and adaptability of symbolic plans for sequential decision making. One instantiation of such framework in \cite{yang:peorl:2018} uses action language $\textbf{BC}$ to formulate dynamic domain through a set of {\em causal laws}, i.e., preconditions and effects of actions and static relationships between properties (fluents) of a state.
In particular, PEORL requires that causal laws formulating cumulative effect (plan quality) defined on a sequence of actions. For an action $a$ executed at state $s$, such causal laws has the form
\[
a~\causes~quality=C+Z~\iif
\]
\[
 ~s,\rho(s,a)=Z,quality=C.
\]
where $\rho$ is a value that will be further updated by reinforcement learning. Reinforcement learning is achieved by R-Learning \cite{rlearning:schwartz,sm:mlj96}, i.e., performing value iteration 
\beq
\ba{rl}
 {R_{t + 1}}({s_t},{a_t})\!\! & \xleftarrow{\alpha_t} {r_t} - {\rho _t}(s_t)  
 + \mathop {\max }\limits_a {R_t}({s_{t + 1}},a),\\
 \rho_{t+1}(s_t)\!\!& \xleftarrow{\beta_t} r_t + \mathop {\max }\limits_a {R_t}({s_{t + 1}},a) - \mathop {\max }\limits_a {R_t}({s_t},a)
\ea
\eeq{rlearning}
to approximate policy that achieves maximal long term average reward using $R$ and gain reward using $\rho$. 

At any time $t$, given an action description in $\textsc{BC}$ and an initial state $I$ and goal state $G$, PEORL uses an answer set solver such as \textsc{clingo} to generate a {\em plan} $\Pi_t$, i.e., a sequence of actions that transits state from $I$ to $G$. After that, the action is sent to execution one by one, value iteration (\ref{rlearning}) is performed. After that, $\rho$ values for all state $s$ in plan $\Pi$ are summed up to obtain the {\em quality of the plan},
$$
quality(\Pi_t)=\sum_{\langle s,a,s'\rangle}\rho(s)
$$
and $\rho(s,a)$ for all $\rho$ values for all transition $\langle s,a,s'\rangle$ are used to update the facts in action descriptions. Plan $\Pi_{t+1}$ is generated that not only satisfies the goal condition $G$, but also has a plan quality greater than the $quality(\Pi_t)$. This process terminates when plan cannot be further improved.

Meta-learning concerns on generating machine learning pipeline with proper hyper-parameter to meet an objective, such as accuracy. This problem can be formulated as an interplay between generating reasonable machine learning pipeline, viewed as a symbolic plan generated from a domain formulation for commonsense knowledge of data science, and evaluating machine learning pipeline, viewed as execution of actions and receiving rewards from the environment, derived from the objective. This approach allows to use interpretable, explicitly represented expert knowledge to delineate search space to look for proper pipeline along with their hyper-parameters, and also allows user to change their specification for the search space in run time, leading to to an more interpretable and transparent meta-learning. The details of the algorithm will be described in Section~\ref{automl}.

\section{Methodology}

\subsection{Knowledge Schema for Data Science}
We first show the knowledge schema defined to capture concepts and relationships in a data science solution. First we define a machine learning interface
{\small
\begin{verbatim}
interface MachineLearningModel {
  id: ID
  algorithm: MachineLearningAlgorithm
  features: [Feature]
  preprocessor: Preprocessor
  saved: Boolean
  accuracy: Float
  timeToLearnInSeconds: Float
  labels: [Label]
 }
\end{verbatim}}
where \texttt{MachineLearningAlgorithm}, \texttt{Feature} and \texttt{Preprocessor} are enumeration type, such as
{\small
\begin{verbatim}
enum MachineLearningAlgorithm {
  random_forest_classifier
  linear_scv_classifier
  gaussian_nb_classifier
  multinomial_nb_classifier
  logistic_classifier
  sgd_classifier
  gradient_boosting_classifier
}
\end{verbatim}}
The interface is implemented by every classifier that is defined as a type that includes an ID and all values for hyper-parameters, for instance
{\small
\begin{verbatim}
type LogisticClassifier 
  implements MachineLearningModel{
  norm: String
  tolerance: Float
  C: Float
  balance: Boolean
  solver: String
  maxIterations: Int
}
\end{verbatim}}

In order to train a classifier, the training input consists of a training data specified by an URL or path. Field inputs consists of user-defined preference for applying featurizers to a column:
{\small
\begin{verbatim}
input FieldInput {
  name: String!
  type: FeatureType!
  featurizerName: [FeaturizerAlgorithm] 
}
\end{verbatim}}
The user can also specify minimal accuracy required for search to stop based on a selection criteria (accuracy, F1, precision, recall), candidate models and candidate preprocessors. The full definition of training input type is 
{\small
\begin{verbatim}
input TrainingInput {
  modelId: ID
  minimumAccuracy: Float
  targetName: String!
  dataInput: TrainingDataInput!
  fields: [FieldInput]
  folds: Int
  selectionCriteria: Metric
  candidateModels: [MachineLearningAlgorithm]
  candidatePreprocessors: [PreprocessorAlgorithm]
  modelProfilingEpisode: Int
  modelSearchEpisode: Int
}}
\end{verbatim}}
The mutation that triggers Meta-learning accepts a training input and outputs an instance of MachineLearningModel.
{\small
\begin{verbatim}
trainClassifier(input: TrainingInput!): 
    MachineLearningModel
\end{verbatim}}
After a machine learning model is trained, it can be used to classify new input data in JSON format:
{\small
\begin{verbatim}
classifyInstances(modelID:ID!,data:JSON!)
\end{verbatim}}
which will return instances of the type \begin{verbatim}[Label]\end{verbatim}.

The schema defined as above can be viewed as taxonomies, or, structure of the data, that will be used to store the results generated during automated machine learning algorithm (Section~\ref{automl}). Encoding the knowledge this way enables the pipeline search process to be better understood, insights about deriving the optimal pipeline to be encoded and inference tasks on Meta-learning to be performed.
\subsection{Automated Machine Learning}\label{automl}
Automated machine learning algorithm is based on PEORL framework, where pipeline generation is viewed as a symbolic planning problem, and pipeline evaluation as reinforcement learning problem that the actions are performed on manipulating data and rewards are derived from cross validation scores. This approach significantly accelerates learning by incorporating domain knowledge to guide exploration, enables online injection of user feedback that changes pipeline search in a flexible and easy way, and interpretable learning in the sense that the learning is performed only on reasonable pipelines generated from a pre-defined symbolic knowledge.
\subsubsection{Representing Domain Knowledge}
We use action language $\mathcal{BC}$ to represent dynamic domain of ML operations and translate the causal laws into corresponding answer set program (ASP). We first introduce three types of objects:
\begin{itemize}
\item Preprocessors, including 
\begin{itemize}
\item matrix decompositions: \\ ({\tt truncatedSVD}, {\tt pca},{\tt kernelPCA}, {\tt fastICA}), 
\item kernel approximation: \\ ({\tt rbfsampler}, {\tt Nystroem}), 
\item feature selection: \\ ({\tt selectkbest}, {\tt selectpercentile}), 
\item scaling: \\ ({\tt minmaxscaler}, {\tt robustscaler}, {\tt absscaler}), and
\item no preprocessing ({\tt noop}),
\end{itemize}
\item Featurizers: \\ including two standard featurizers for text classification, i.e., {\tt CountVectorizor} and {\tt TfidfVectorizer}.
\item Classifiers: \\ including logistic regression, Gaussian naive Bayes, linear SVM, random forest, multinomial naive Bayes and stochastic gradient descent.
\end{itemize}

We treat each operation in the pipeline as an action, with describing their causal laws accordingly. First of all, it includes facts about compatibility with sparse vectors, such as
{\small
\begin{verbatim}
acceptsparse(random_forest_classifier)
\end{verbatim}}
facts about operators and data type, such as
{\small
\begin{verbatim}
compatible(integer,std_scaler)...
\end{verbatim}}
and actions such as data import, train and predict. The actions related to configuration of machine learning pipelines are described as follows:
\begin{itemize}
\item Select featurizers for each column. The following rules describes the effects of initialize featurizer for each column, and after each column has one featurizer selected, the cumulative quality increments. 
{\small
\begin{verbatim}
{initfeaturizer(F,Field,Y,k)}:-
  feature(F),
  has_attr(Field,Y,k),
  compatible(Type,F),
  has_type(Field,Type), 
  datatype(Y,train),
  not
    modeltrained(Y,k), 
  not
    featurizerinitialized(F,Field,Y,k).
  
featurizerinitialized(F,Field,Y,k):- 
    initfeaturizer(F,Field,Y,k-1),
    feature(F).
cost(Q+R,k) :-
    featurizationcompleted(k),
    ro(R,k),cost(Q,k-1).
cost(Q+10,k):-
    featurizationcompleted(k), 
    #count{R:ro(R,k)}=0, cost(Q,k-1).
featurizationcompleted(k):- 
    #count{Field:featurizerinitialized(_,
    Field,_,k)}=X1,
    #count{
        Field:has_field(_,Field)}=X2,
        X1=X2.
\end{verbatim}}
\item Select preprocessor:
{\small
\begin{verbatim}
{initpreprocessor(P,Y,k)}:- 
    featurizationcompleted(k),
    preprocessor(P), 
    not 
      modeltrained(Y,k),
    datatype(Y,train),
    featurizationcompleted(k).
preprocessorinitialized(P,Y,k) :- 
    initpreprocessor(P,Y,k-1),
    featurizationcompleted(k).
cost(Q+R,k) :-
    initpreprocessor(P,Y,k-1),
    ro(P,R,k-1),
    cost(Q,k-1).
cost(Q+10,k) :- initpreprocessor(P,Y,k-1), 
    #count{R:ro(P,R,k-1)}=0,
    cost(Q,k-1).
\end{verbatim}}
\item Crossvalidate. If we have token and label from the data, we can cross validate the pipeline by choosing featurizers, preprocessor and a classifier, and the effect is the model being validated. If one of preprocessor and classifier does not accept sparse vector, it needs to be transformed into dense vector. 
{\small
\begin{verbatim}
sparse :- has_type(X,text), 
    #count{T:has_type(X1,T)}=1.
{crossvalidate(C,P,dense,T,k)} :-
    classifier(C), 
    preprocessorinitialized(P,Y,k),
    has_attr(T,Y,k),
    not
      sparse,
    has_targetfield(data,T).
{crossvalidate(C,P,sparse,T,k)} :- 
    classifier(C),
    has_attr(T,Y,k), 
    preprocessorinitialized(P,Y,k),
    acceptsparse(P),
    acceptsparse(C),
    sparse.
modelvalidated(C,P,S,T,k) :-
    datatype(Y,train),
    crossvalidate(C,P,S,T,k-1).
modelvalidated(T,k) :-
    datatype(Y,train),
    crossvalidate(C,P,S,T,k-1).
cost(Q+R,k) :-
    ro(P,C,R,k-1),
    cost(Q,k-1),
    crossvalidate(C,P,S,T,k-1).
cost(Q+10,k) :-
    crossvalidate(C,P,S,T,k-1), 
    #count{R:ro(P,C,R,k-1)}=0,
    cost(Q,k-1).
\end{verbatim}}
\end{itemize}

Besides causal laws described above, all fluents are declared inertial, and concurrent execution of actions are prohibited except for {\tt initfeaturizer}.

Furthermore, to facilitate fast profiling, we use empirical assignment of featurizers to data types, unless they are overriden by the user. This capability is enabled leveraging the non-monotonic reasoning capability of answer set programming. The default application of featurizers are: for categorical data type, apply \texttt{one\_hot}, for flat data type, apply \texttt{std\_scaler}, for integer type, apply \texttt{min\_max\_scaler}, for text, apply \texttt{hashing\_vectorizer} (bag of words). One example of these rules is
{\small
\begin{verbatim}
:- initfeaturizer(F,Field,Y,k), 
    has_type(Field,categorical),
    F!=one_hot
    #count{F1:use_featurizer(Field,F1)}=0.
\end{verbatim}}
\subsubsection{Generation of Pipeline}

The generation of pipeline is treated as a symbolic planning problem given the above formulation of the dynamic domain. The initial condition comes from two sources: (1) data schema and (2) configuration of pipeline search space. The data schema consists of the column name and its data type extracted from a data source (e.g., a CSV file in which each column has been pre-labelled with their data type and one column is designated as classification target). For instance, the following ASP file is generated for Adult\footnote{https://archive.ics.uci.edu/ml/datasets/Adult} dataset, where {\tt field\_salary} is the classification target:
{\small
\begin{verbatim}
datatype(adult_data,train).
datatype(adult_test,test).
has_field(data,field_age).
has_type(field_age,integer)...
has_targetfield(adult_data, field_salary).
\end{verbatim}}
The configuration of pipeline search space comes from user specification on candidate featurizers, candidate preprocessors ad candidate models. This information is translated into ASP file as well. For example, the following file configure the pipeline search to be performed amoung the listed classifier, preprocessors and featurizers:
{\small
\begin{verbatim}
classifier(linear_svc_classifier)...
feature(one_hot)...
preprocessor(pca)...
\end{verbatim}}
Furthermore, user can override default application choices of featurizers by specifying their own preference. For instance, user wants to use \texttt{robustscaler} for column \texttt{Age}, then the following fact is appended to the ASP file:
{\small
\begin{verbatim}
use_featurizer(robust_scaler,field_age).
\end{verbatim}}
The planning goal to train a classifier is defined to be
{\small
\begin{verbatim}
:- not modeltrained(adult_data,k), query(k).
:- query(k),cost(Q,k), Q<=1.
\end{verbatim}}
Alternatively, given a test file, the goal can also be classifying the test data, with the first constraint replaced by
{\small
\begin{verbatim}
:- not has_attr(adult_test, field_salary,k).
\end{verbatim}}

Given the initial condition and a goal, plan can be generated by translating the action description above to ASP and run answer set solver {\sc clingo}:
{\small
\begin{verbatim}
1:import_train(adult_data)
2:initfeaturizer(one_hot,field_sex,
    adult_data) 
  initfeaturizer(one_hot,field_race,
    adult_data) 
  initfeaturizer(one_hot,field_education,
    adult_data) 
  initfeaturizer(one_hot,field_workclass,
    adult_data) 
  initfeaturizer(robust_scaler,field_age,
    adult_data)
3:initpreprocessor(Nystroem,adult_data)
4:crossvalidate(gradient_boosting_classifier,
    nystroem,dense,field_salary)
5:train(gradient_boosting_classifier,
    nystroem,dense,field_salary) 
\end{verbatim}}
The above output is a plan that achieves the goal from the initial state. The machine learning pipeline is encoded into operations step by step, including initializing the featurizer in step 2, initializing the preprocessor in step 3, and picking up a classifier to perform cross validation in step 4. In practice, the plan request is sent from a front-end UI or a GraphQL query, where there are other hyper-parameters that need to be specified, which will be described in Section~\ref{system}.


Currently we only allow one featurizer to be applied to a column. In the future, we will allow user to specify multiple featurizers to be applied to a column, to enable more flexible feature specification and increase the expressivity of search space of pipelines.

\begin{algorithm}
{\small
  \caption{Pipeline Profiling}
  \label{algexec}
  \begin{flushleft}
    \begin{algorithmic}[1]
        \REQUIRE candidate classifiers $\mathbb{C}$, candidate preprocessors $\mathbb{P}$ and user-specified featurizer-column application $\mathbb{F}$, planning goal, $G=(\texttt{:- not~modeltrained(\i{data})},\emptyset)$, domain representation $D$, profiling episodes $\i{profilingepisodes}$ and cross-validation folds $v$.
        \STATE $P_0\Leftarrow \emptyset$, $\Pi\Leftarrow \emptyset$
        \STATE generate initial state $I$ from $\mathbb{C}$, $\mathbb{P}$, $\mathbb{F}$.
        \WHILE{True}
          \STATE $\Pi_o\Leftarrow \Pi$
          \STATE solve planning problem $\Pi\Leftarrow\textsc{Clingo}.\i{solve}(I,G,D\cup P_t)$
          \IF {$\Pi=\Pi_o$}
              \RETURN $\Pi_o$
          \ENDIF
          \FOR {action $\langle s,a,s'\rangle \in\Pi$}
          \IF {$a\in\{\i{initfeaturizer}(F,Col,Y)$, $\i{initpreprocessor}(P,Col,Y)\}$ where $F\in\mathbb{F}$,$P\in\mathbb{P}$, $\i{Col}$ is a column name and $Y$ is the training dataset name}
            \STATE $\i{reward}\Leftarrow -1$
          	\STATE update $R(a,s)$ and $\rho_t^a(s)$ for action $a$
          \ENDIF
          \IF {$a\in\{\i{crossvalidate}(C,P,D,Col,Y)\}$ where $C\in\mathbb{C}$, $\i{Col}$ is a column name and $Y$ is the training dataset name} 	
            \FOR {$i<\i{profilingepisode}$}
            	\STATE instantiate $C,P,F$ by random sampling their hyper-parameters.
            	\STATE assemble pipeline using $C,P,F$.
            	\STATE perform $v$-fold cross validation using $C,F,P$.
            	\STATE obtain $\i{reward}\propto \i{cvscore}$.
                \STATE update $R(a,s)$ and $\rho_t^a(s)$ for action $a$
                \STATE $i\leftarrow i+1$
            \ENDFOR
          \ELSE
          \STATE execute $a$
          \ENDIF
          \ENDFOR
          \STATE $quality(\Pi)\leftarrow \sum_{\langle s,a,s'\rangle\in\Pi}\rho^a(s)$
          \STATE update planning goal $G\Leftarrow (A, quality> quality(\Pi))$.
          \STATE update facts $P_t\Leftarrow \{\rho(a)=z:\rho_t^{a}(s)=z\}$.
        \ENDWHILE
    \end{algorithmic}
    \end{flushleft}}
\end{algorithm}

\subsubsection{Pipeline Learning}
The evaluation of pipeline is performed using the algorithm based on PEORL, shown in Algorithm~\ref{algexec}. The algorithm accepts a set of candidate classifiers, a set of candidate preprocessors and user-specified application of featurizers to columns and a training dataset. The input parameter to the data also consists of model profiling episodes and cross validation folds. First of all,the initial state of planning problem (line 2) and a pipeline is generated (line 5). The action in the pipeline is executed through calling a library of methods that wraps \textsc{scikit-learn} libraries, and for featurization and preprocessing, give reward -1 to promote shorter pipeline (line 11) and perform value iteration of R-learning (line 12). For cross-validation action, picking up a random parameter for preprocessor, featurizers and classifiers (line 16) to assemble a pipeline (line 17) and perform cross validation (line 18). It involves converting the feature matrix to dense if specified. After that, deriving reward proportional to the average cross validation score (line 19) and update $R$ and $\rho$ values using R-Learning. After the whole pipeline is profiled, evaluating the pipeline quality (line 27), update planning goal (line 28) and write back $\rho$ values into symbolic formulation to generate a new pipeline. When the pipeline cannot be further improved, return the optimal pipeline (line 6). In our application pipeline output takes the form of a serial pickle object.

\subsubsection{Interactive Process of Meta-Learning}

Meta-Learning is by its nature a long running job, especially with larger datasets. The PEORL algorithm applied to machine learning by itself does not scale well. Because of this we have incorporated two features with the intention of improving performance. First, before running the PEORL algorithm, we have a broad search phase were we test each classifier once. We then restrict the algorithm to only the best performing classifier and continue the search process.

\begin{enumerate}
	\item \textbf{Model Selection} (Phase 1). For $\mathbb{C}={C}$ for each classifier $C$, call Algorithm~\ref{algexec} using the chosen feature set, and $\mathbb{P}=\{\i{noop}\}$, recording the performance. Select classifier $C_0$ based on the predefined selection criteria (accuracy, F1, precision, recall).
	\item \textbf{Pipeline Learning} (Phase 2).  Call Algorithm~\ref{algexec} with $\mathbb{C}=\{C_0\}$, $\mathbb{P}$ and $\mathbb{F}$ being user specified featurizers and preprocessors, and generate optimal pipeline $\Pi_1$
	\item \textbf{Parameter Sweeping.} (Phase 3). Perform grid search or random search for hyper parameter $\Pi_1$ and return the final pipeline $\Pi_2$.
\end{enumerate}

We also allow the user to gradually refine their preference during the process. When the user sees the results, they may inject their feedback by overriding any preset configurations above, at any time, and this information will be picked up by Meta-Learning search algorithm, and change its behavior towards user's feedback in the next episode of planning and learning. The user can remove or add possible algorithms to test, cancel the current pipeline, stop a phase with the current best classifier, or stop the entire process and use the best classifier found.

\section{System Implementation}\label{system}
\begin{figure*}
\begin{subfigure}{0.49\textwidth}
    \centering
    \includegraphics[scale=0.4]{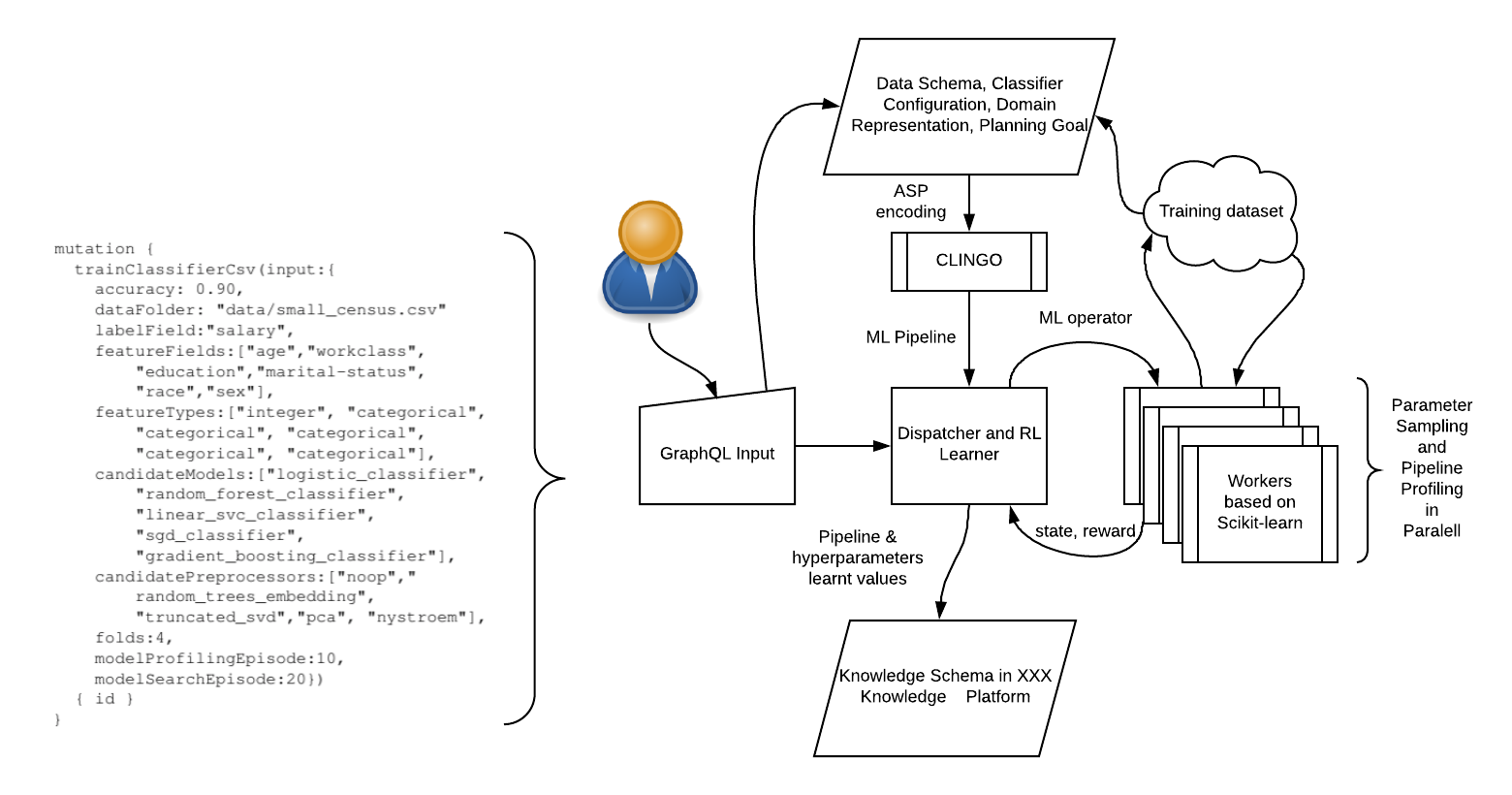}
    \caption{The Meta-learning Service Architecture.}
    \label{fig:architecture}
\end{subfigure}
\begin{subfigure}{0.49\textwidth}
    \centering
    \includegraphics[scale=0.25]{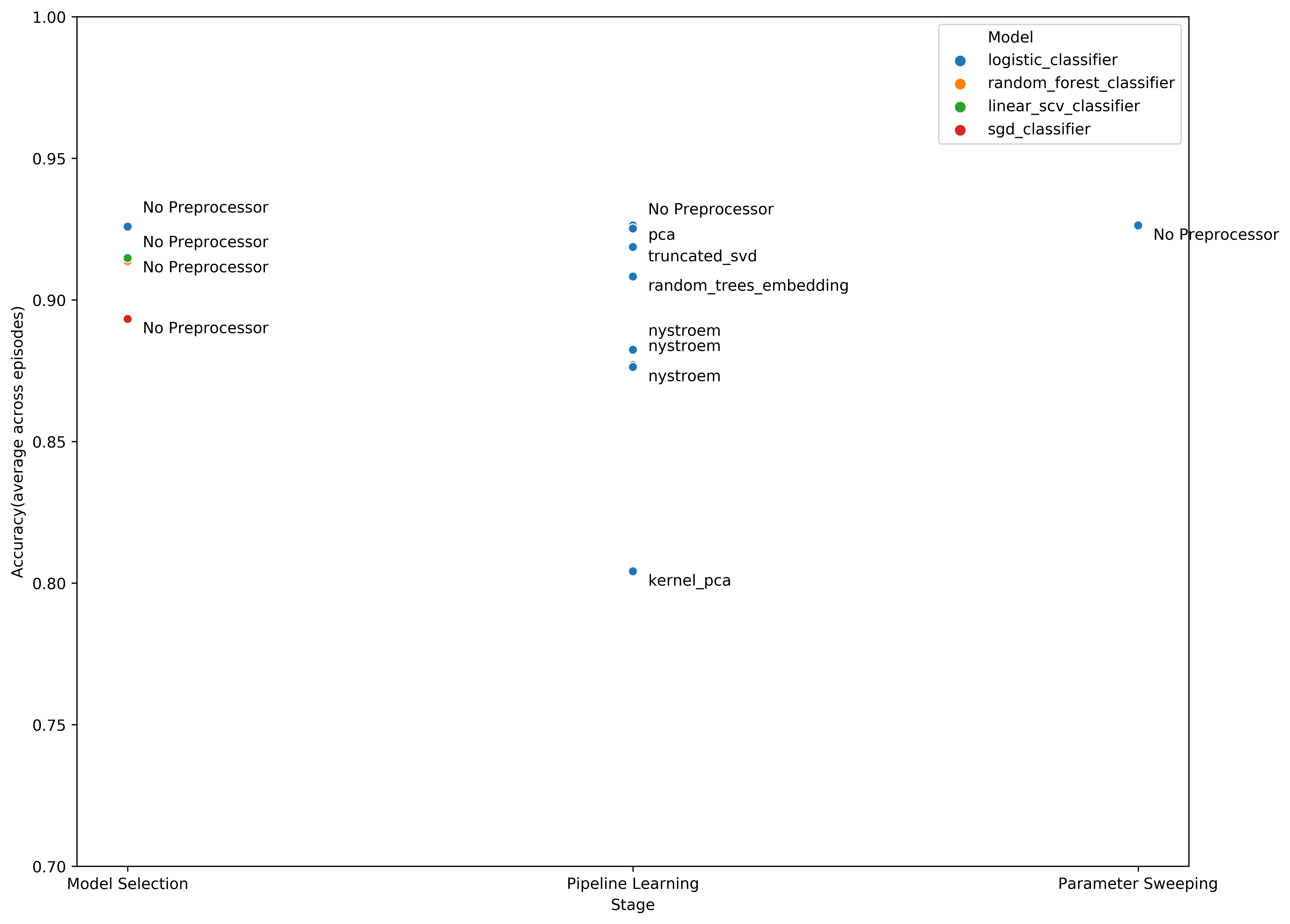}
    \caption{Results of the three stages.}
    \label{fig:modelselection}
\end{subfigure}
\begin{subfigure}{1\textwidth}
    \centering
    \includegraphics[scale=0.4]{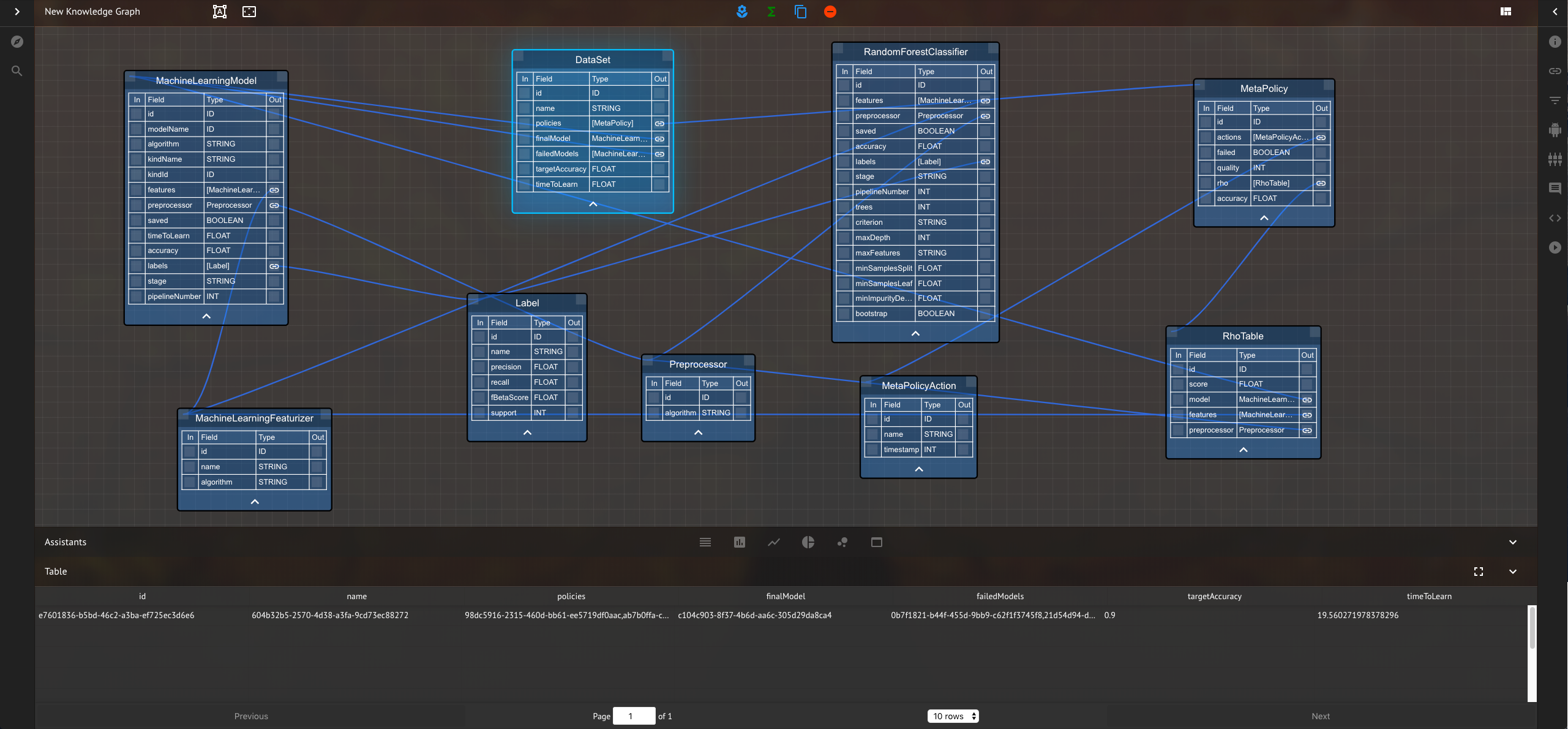}
    \caption{The results in the Knowledge Graph.}
    \label{fig:kg}
\end{subfigure}
\caption{Meta-learning Service Overview}
\end{figure*}

\begin{table*}[!thb]
    \centering
    \begin{tabular}{|c|c|c|c|c|}
    \hline
    Dataset & Featurizer & Preprocessor & Classifier & CV accuracy \\
    \hline\hline
    Reuters 50/50 & hashing vectorizer & none & linear SVC & 0.848 \\
    IMDB & hashing vectorizer & None & SGD classifier & 0.879 \\
    Adult & one hot, min\_max scaling & random trees embedding & logistic regression & 0.8523 \\
    Spam detection & min\_max scaling & none & logistic regression & 0.927 \\
    Parkinsons detection & std\_scaler & none & random forest & 0.887 \\
    Abalone & std\_scaler & Nystroem & random forest & 0.552\\
    Car & one hot & Nystroem & gradient boosting & 0.938\\
    \hline
    \end{tabular}
    \caption{Baseline Pipelines Learned on Datasets}
    \label{tab:my_label}
\end{table*}

In Maana Knowledge Platform, CSV files can be uploaded and each column becomes a field and their types are automatically identified. The user can trigger Meta-learning service by submitting a query through GraphQL endpoint. The GraphQL input is used to generate part of the initial state for planning, and the Meta-learning service is triggered for pipeline search. Throughout the pipeline search process, the results are constantly written to the Maana Knowledge platform according to the knowledge schema. The service is implemented in Python with Graphene library to enable GraphQL server and endpoints. It is deployed using a Docker image along with other components of Maana Knowledge platform.

Additionally, another feature we use to improve performance is the parallelization of building models. Because the model profiling and model search episodes can be done in parallel, we use asynchronous approach, where multiple workers are launched and each perform their own parameter sampling and cross validation on the dataset, and the result is returned to the dispatcher to perform value iteration.

\subsection{Example: Classify Spam Email}\label{sec:example}
We show how the Meta-learning service perform on an example dataset obtained from UCI machine learning data repository: Spam message detaction\footnote{https://archive.ics.uci.edu/ml/datasets/Spambase}. The dataset contains 4601 data entries, 57 float and integer features to detect if a message is a spam or not.

After Meta-learning service is running, the user can load a CSV file into a workspace into Maana project. After that, the user launch the service through the GraphQL endpoint, where the user specifies feature fields and their related types, candidate classifiers (logistic regression, random forest, linear SVC, SGD classifier) and candidate preprocessors (noop, random trees embedding, truncated SVD, PCA, Nystroem, kernel PCA). It performs 10 folds cross validation, 10 episodes of model profiling and 20 episodes of model search.

After the service is launched, it goes to the first phase, model selection. In this phase, it will not apply any preprocessors and only apply default featurizers to the column, and pick up 10 sets of random hyper parameters for each model and calculate the average cross validation accuracy. By the end of model selection (Phase 1), it stores the metric information into the knowledge graph, and is visualized in the first column of Fig.~\ref{fig:modelselection}. It shows that the most accurate model, based on cross validation results, is logistic regression. At this point, the user is notified that logistic regression is selected, based on predefined selection criteria. In Pipeline Learning (Phase 2), Meta-learning will try to find the best combination of preprocessors and featurizers using the selected classifier, following Algorithm~\ref{algexec}. Since the user does not override any default selection of featurizers, \texttt{one\_hot\_encoder} is applied to categorical fields, and \texttt{min\_max\_scaler} is applied to integer fields. During this process, ASP-based planner generates pipelines using the selected classifier, and reinforcement learner evaluates the generated pipeline on the data, using reward derived from the cross validation accuracy. The pipeline is gradually improved till the point that it does not change. By the end of Phase 2, the performance of selecting different preprocessor with the classifier is output in Fig.~\ref{fig:modelselection}. From the result, it shows not performing any preprocessor has the best performance used with logistic regression. Combined with the default featurizer, a pipeline of using min\_max scaler for integer fields, one hot encoder for categorical field, random tree embedding for preprocessor and a logistic regression is learned. Finally, during parameter sweeping (Phase 3) hyper-parameters are swept, leading to the final results in the third column. All of the intermediate search results are stored in the knowledge graph, shown as a snapshot in Fig.~\ref{fig:kg}. The upper part of the screen shots shows the knowledge schema organized as knowledge graph, and on clicking each of the schema node, data instance is shown in the lower part of the workspace.

During this process, based on the pre-defined candidate models, preprocessors and the profiling episodes, the system has evaluated 4 pipelines (each parameterized with 10 sets of hyper parameters) in model Selection process. During pipeline learning phase, the 8 pipelines based on selected classifier (logistic regression) and candidate preprocessors are further evaluated (with 10 hyper-parameter tested in a single learning episode) until the optimal pipeline converges. This process does not provide systematic pipeline optimization and search. Instead, it leverages the decision space pre-defined by the data scientist and perform quick profiling and provide evidence for the data scientist to further refine their decisions.

\subsection{Evaluation on Datasets}

We evaluate meta-learning services for classification tasks using default setting for featurizers. Our data set are obtained includes:
\begin{itemize}
\item Reuters 50/50 dataset\footnote{https://archive.ics.uci.edu/ml/datasets/Reuter\_50\_50} contains of 2,500 texts (50 per author) for author identification.
\item IDMB movie review dataset\footnote{http://ai.stanford.edu/~amaas/data/sentiment/} contains 25,000 movie reviews obtained from IMDB. The classification task is to predict a movie review is positive or negative.
\item Adult dataset\footnote{https://archive.ics.uci.edu/ml/datasets/Adult} contains 48842 instances. Each instance has 14 fields, including age (integer), working class (categorical), education (categorical), capital gain (float), etc that consitute the feature space to predict one of the two classs: salary $>50k$ or $<=50k$.
\item Spam email detection\footnote{https://archive.ics.uci.edu/ml/datasets/Spambase} contains 4601 data entries, 57 float and integer features to detect if a message is a spam or not.
\item Parkinson's detection dataset \footnote{https://archive.ics.uci.edu/ml/datasets/Parkinsons} contains 197 instances, with 22 float features to detect if a person has Parkinson's disease or not based on vocal characteristics.
\item Abalone dataset\footnote{https://archive.ics.uci.edu/ml/datasets/Abalone} contains 4177 instances, 8 float and integer attributes to detect the sex of abalone.
\item Car evaluation dataset\footnote{https://archive.ics.uci.edu/ml/datasets/Car+Evaluation} contains 1728 instances, 6 categorical data fields to make classification of purchasing decisions.
\end{itemize}
Detailed results are shown in Table~\ref{tab:my_label}. The result shows that the Meta-learning service generates competitive baseline result for the data scientist to further work on.

\section{Conclusion and Future Work}

Meta-learning service provides a novel framework for machine learning pipeline search that is transparent, interpretable and interactive. It serves as a profiling tool for data scientist to use: by incorporating human knowledge, meta learning service performs efficient pipeline generation and profiling in the search space delineated by the data scientists, allowing feedback to be injected in the middle to alter search space and providing useful feedback for the data scientist to understand the best machine learning pipeline for the dataset of interest. While the Meta-learning service will by no means replace data scientist to finish data science project automatically, it can save a large amount of time for manual search and tuning. Currently it is deployed in Maana Knowledge Platform to facilitate data scientists to build machine learning solutions faster, with better insight and facilitate knowledge management and sharing across different projects.

This framework leaves several paths for improvement. Up until now we have only applied featurization to data based on what type it is. However, it is possible to perform some level of automated feature extraction and clean up of the data. It should be possible to use the knowledge gleaned from the meta-attributes to guide the algorithm as well. More advanced parameter optimization can also be applied.

\bibliographystyle{IEEEtran}
\balance  
\bibliography{root}
\end{document}